\documentclass{article}

\usepackage[sglblindworkshop, final]{neurips_2026}
\workshoptitle{SAPC2 Competition Workshop at NeurIPS 2026}

\usepackage[utf8]{inputenc}
\usepackage[T1]{fontenc}
\usepackage{hyperref}
\usepackage{url}
\usepackage{booktabs}
\usepackage{amsfonts}
\usepackage{nicefrac}
\usepackage{microtype}
\usepackage{xcolor}
\usepackage{graphicx}

\title{Fine-Tuning a Transducer for Impaired Speech:\\ Where the Gains Appear and\\ What They Mean for Evaluation}

\author{%
  Kien Nguyen \qquad Tuan Dinh \qquad Bao Phan \\
  Eloky \\
  \texttt{kien@eloky.ai} \qquad \texttt{tuanad121@gmail.com} \qquad \texttt{bao@eloky.ai}
}

\begin{document}

\maketitle

\begin{abstract}
Speech recognition still works poorly for many people whose speech is affected by conditions such as ALS, Parkinson's
disease, cerebral palsy, Down syndrome or stroke. We took part in the Speech Accessibility Project Challenge~2 (SAPC2)
with a public speech recogniser that we fine-tuned on the challenge data. Our best entry scored 7.81\%
character error rate on Test1, the first of two hidden test sets (official baseline 15.26\%). Our live entries
ran on CPUs with a combined median latency of about 0.9 seconds at 7.96--8.07\% error. On new speakers,
fine-tuning reduced errors most for the speakers that an off-the-shelf recogniser finds hardest to understand.
Sentences already seen in training had far lower error rates than new ones (2.58\% vs 6.42\%), a
larger difference than being a familiar speaker made. We close with lessons for evaluating such systems.
\end{abstract}

\section{Introduction}

Voice interfaces are now everywhere, yet they make far more mistakes for people with speech impairments, who may need
them most. A leaderboard tells us which system won, but not what a team building a product for these users needs to
know. We use our SAPC2 entry to study four questions:
\begin{enumerate}\setlength\itemsep{0pt}
  \item How much does fine-tuning a public model reduce errors, for which conditions and at which severity?
  \item How do familiar speakers and familiar sentences relate to accuracy?
  \item Can models of this kind serve both offline and live use on ordinary computers?
  \item What general lessons did we learn about evaluating improvements?
\end{enumerate}
We report \emph{character error rate} (CER): the share of characters that are wrong after the official text
normalisation. Lower is better. Tables also give \emph{word error rate} (WER), the same measure counted in words.

\textbf{Background.} Early collections of impaired speech, such as UASpeech~\citep{kim2008uaspeech} and
TORGO~\citep{rudzicz2012torgo}, are small and mostly contain isolated words. Project
Euphonia~\citep{macdonald2021euphonia} and the Speech Accessibility Project (SAP)~\citep{hasegawajohnson2024sap}
collected speech from hundreds to thousands of people. SAP's recordings come from people with Parkinson's disease, ALS,
cerebral palsy, Down syndrome and stroke, and its challenges~\citep{zheng2025sapc,sapc2website} measure accuracy on
speakers who are not in the training data. SAPC2 has two tracks: Track~1 scores offline transcription, and Track~2
scores live (streaming) transcription on both accuracy and speed.

\section{Data and how we measured}
\label{sec:method}

\textbf{Data.} Each SAP recording is one prompt: a short command, a read sentence or an answer to a question. As the
challenge allows~\citep{sapc2website}, we trained on the challenge training set together with the training and
development parts of the SAP research data release: about 950 hours from over 1{,}100 speakers. Our later entries also
trained on the challenge's labelled development speakers. We used no speech from outside SAP and no synthetic speech.
The organisers' published speaker counts for the hidden test sets (both combined) are ALS~66, cerebral palsy~58, Down
syndrome~32, stroke~24 and Parkinson's disease~13~\citep{sapc2website}.

\textbf{Evaluation sets.} We split the challenge's labelled development set by speaker and used four sets:
\begin{itemize}\setlength\itemsep{0pt}
  \item \textbf{Selection panel}: 41 speakers, 4{,}510 recordings, examined only through aggregate scores.
  \item \textbf{Former sealed group}: 83 speakers, kept sealed at first and later added to training.
  \item \textbf{Held-back recordings} of 300 training speakers, used only for the familiarity study
  (Section~\ref{sec:familiarity}).
  \item \textbf{Test1}: the official hidden test set. Entries trained on all labelled speakers can only be judged here.
\end{itemize}
The severity and familiarity analyses use all labelled recordings of the 124 new speakers (47,929 recordings),
including the many extra recordings of repeated prompts in the challenge's development data, so most of these
recordings are sentences that also appear in training. All scores use the official scoring tool. The ranges in our
tables are 95\% intervals from resampling \emph{speakers} 2{,}000 times (``speaker intervals''), because a product
meets new speakers, not just new recordings. We never accessed the second hidden test set and never adapted to test
audio.

\section{Our system in brief}

We kept the system simple and built it from public parts.
\begin{itemize}\setlength\itemsep{0pt}
  \item \textbf{Starting model.} We compared several public recognisers and chose Parakeet TDT 1.1B~\citep{parakeet11b,
  rekesh2023fastconformer,xu2023tdt}, a 1.1-billion-parameter model whose decoder also predicts how long each token
  lasts (``TDT'').
  \item \textbf{Fine-tuning.} We fine-tuned the whole model on the impaired-speech data with NeMo~\citep{kuchaiev2019nemo}.
  \item \textbf{Averaging.} We averaged the weights of fine-tuned models~\citep{wortsman2022soups}. The result is one
  model, so this adds no cost when the system runs.
  \item \textbf{Live use.} Our live entries run fine-tuned models of the same kind on CPUs.
\end{itemize}

\section{Results}

\subsection{Overall accuracy and accuracy by condition}

\textbf{Hidden test set.} Table~\ref{tab:test1} shows our official Track~1 scores. Our best entry trained without the
selection speakers has 7.87\% CER, 48\% fewer character errors than the official
baseline. Our best scored entry, trained on all labelled speakers, has 7.81\%.

\begin{table}[t]
  \centering
  \caption{Official Track~1 results on the hidden Test1 set (\%). Results for the second test set were not released
  when this version was written.}
  \label{tab:test1}
  \small
  \begin{tabular}{lccc}
\toprule
System & Trained on selection speakers & CER & WER \\
\midrule
Official challenge baseline (organisers) & -- & 15.26 & 21.53 \\
Earlier entry & no & 8.30 & 11.95 \\
Best entry trained without the selection speakers & no & 7.87 & 11.78 \\
Best scored entry, trained on all labelled speakers & yes & 7.81 & 11.69 \\
\bottomrule
\end{tabular}
\end{table}

\textbf{By condition.} Table~\ref{tab:etiology} compares our averaged model with two recognisers that were not
fine-tuned on this data: the original Parakeet model, and Whisper large-v3~\citep{radford2023whisper,fasterwhisper} as
used in the challenge starter kit. Our column is the model of our best entry trained without the selection speakers.
On the selection panel, it has about half the errors of either, and lower point estimates for every condition, although
the intervals overlap for Parkinson's disease, ALS and stroke. Parkinson's disease and ALS have the lowest error rates; cerebral palsy, Down syndrome and
stroke remain harder.

\begin{table}[t]
  \centering
  \caption{CER (\%) on the selection panel (41 new speakers) by condition, with 95\% speaker intervals.
  ``Ours'' is the model of our best entry trained without the selection speakers. $^\dagger$Only four speakers.}
  \label{tab:etiology}
  \small
  \setlength{\tabcolsep}{3pt}
  \resizebox{\linewidth}{!}{\begin{tabular}{lrrrrr}
\toprule
Condition & Speakers & Recordings & Ours (averaged) & Parakeet TDT 1.1B & Whisper large-v3 \\
& & & fine-tuned & before fine-tuning & off the shelf \\
\midrule
Parkinson's disease & 11 & 1,356 & 4.76 {\scriptsize [2.37, 7.47]} & 7.51 {\scriptsize [3.21, 13.29]} & 8.00 {\scriptsize [3.72, 13.18]} \\
ALS & 10 & 522 & 4.95 {\scriptsize [2.02, 11.67]} & 8.60 {\scriptsize [3.82, 19.38]} & 11.60 {\scriptsize [6.32, 21.39]} \\
Cerebral palsy & 9 & 1,058 & 8.49 {\scriptsize [5.01, 14.34]} & 20.37 {\scriptsize [10.07, 40.11]} & 19.59 {\scriptsize [11.88, 30.79]} \\
Down syndrome & 7 & 1,066 & 7.00 {\scriptsize [5.01, 9.72]} & 14.73 {\scriptsize [10.12, 19.74]} & 15.12 {\scriptsize [10.46, 20.65]} \\
Stroke$^\dagger$ & 4 & 508 & 8.56 {\scriptsize [3.00, 17.21]} & 20.59 {\scriptsize [7.44, 50.68]} & 20.76 {\scriptsize [6.83, 41.66]} \\
\midrule
All & 41 & 4,510 & 6.09 {\scriptsize [4.52, 8.00]} & 12.16 {\scriptsize [8.44, 16.75]} & 12.73 {\scriptsize [9.44, 16.48]} \\
\bottomrule
\end{tabular}}
\end{table}

\textbf{Test1 versus the selection panel.} For that entry, Test1 CER (7.87\%) was about
1.29 times its selection-panel CER.
Reweighting the panel's conditions by the published hidden-set speaker counts raises its CER from 6.09\% to
6.79\%, suggesting that the mix of conditions may explain part of the difference.

\subsection{Accuracy by severity}

We grouped the 124 new speakers by how well an off-the-shelf recogniser (Whisper) understands them, and scored every
model on the same speakers and recordings. This grouping is not a clinical severity rating. We compare against the
\emph{other} recogniser, the original Parakeet model before fine-tuning, since grouping by Whisper's own errors would
bias Whisper's numbers. ``Fine-tuned'' is the single fine-tuned analysis model of Section~\ref{sec:familiarity}, not
our best entry.

\begin{table}[t]
  \centering
  \caption{CER (\%) of the 124 new speakers grouped by their CER under Whisper large-v3, with 95\% speaker intervals.
  Groups are Whisper CER ranges in \%. ``Fine-tuned'' is the analysis model; ``Original'' is Parakeet before
  fine-tuning; ``Removed'' is the share of its errors that fine-tuning removed. ``New sentences only'' keeps recordings whose text is not in
  the analysis model's training data; groups are recomputed on those recordings, so speaker counts differ.}
  \label{tab:severity}
  \small
  \setlength{\tabcolsep}{3pt}
  \resizebox{\linewidth}{!}{\begin{tabular}{lrllrrll}
\toprule
& \multicolumn{4}{c}{New sentences only} & \multicolumn{3}{c}{All recordings} \\
Group (Whisper CER) & Spk. & Fine-tuned & Original & Removed & Spk. & Fine-tuned & Original \\
\midrule
Mild ($<$5) & 37 & \textbf{1.91} {\scriptsize [1.53, 2.30]} & 2.38 {\scriptsize [1.99, 2.77]} & 20\% & 41 & \textbf{1.22} & 2.76 \\
Moderate (5--15) & 38 & \textbf{4.04} {\scriptsize [3.15, 5.05]} & 5.57 {\scriptsize [4.46, 6.85]} & 27\% & 34 & \textbf{2.55} & 6.49 \\
Severe (15--30) & 23 & \textbf{10.89} {\scriptsize [8.93, 13.37]} & 17.00 {\scriptsize [11.96, 23.86]} & 36\% & 22 & \textbf{7.09} & 18.08 \\
Very severe ($\geq$30) & 26 & \textbf{30.62} {\scriptsize [24.98, 37.46]} & 51.88 {\scriptsize [41.62, 62.15]} & 41\% & 27 & \textbf{16.38} & 56.44 \\
\bottomrule
\end{tabular}}
\end{table}

Table~\ref{tab:severity} leads with new sentences, the cleaner comparison. There, the share of errors removed by
fine-tuning grows with severity: 20\% in the mildest group, 27\% and 36\% in the middle groups, and 41\% in the most
severe, where CER falls from 51.9\% to 30.6\%. In the mildest group the intervals overlap. Over all recordings, which
are mostly known sentences, the original model's error rate rises from 2.8\% to 56\% across the groups, while the
fine-tuned model's rises from 1.2\% to 16\%.

\subsection{Familiar speakers and familiar sentences}
\label{sec:familiarity}

SAP prompts repeat across speakers, so a test recording may share its speaker, its sentence, or both with training
data. To study this, we trained one analysis model with a single fine-tuning run while holding back some recordings of
300 training speakers. This gives four groups: (1)~known speaker, known sentence; (2)~known speaker, new
sentence; (3)~new speaker, known sentence; (4)~new speaker, new sentence. Group~1 is further split into (1a)~the same
person said this sentence in training and (1b)~only other people did. The new speakers are the 124 from the selection
panel and the former sealed group. A sentence counts as known if its text, ignoring case and punctuation, appears in
training. Each speaker can appear in several groups.

\begin{table}[t]
  \centering
  \caption{CER (\%) by group, with 95\% speaker intervals. Groups are labelled speaker / sentence. Group~1 combines
  1a (the same speaker said the sentence in training) and 1b (only other speakers did). ``Fine-tuned'' is the analysis
  model; ``Parakeet'' is Parakeet TDT 1.1B before fine-tuning; ``Whisper'' is Whisper large-v3 off the shelf. ``Known'' refers to the
  analysis model's training data; for the other two models these are simply the same recordings.}
  \label{tab:cells}
  \small
  \setlength{\tabcolsep}{3pt}
  \resizebox{\linewidth}{!}{\begin{tabular}{lrrlll}
\toprule
Group (speaker / sentence) & Rec. & Spk. & Fine-tuned & Parakeet & Whisper \\
& & & (analysis) & original & off the shelf \\
\midrule
1: known / known & 2,362 & 300 & \textbf{1.45} {\scriptsize [1.05, 1.87]} & 13.87 {\scriptsize [11.84, 16.11]} & 13.07 {\scriptsize [11.07, 15.19]} \\
\quad 1a: same speaker & 1,105 & 293 & \textbf{0.75} {\scriptsize [0.47, 1.08]} & 12.90 {\scriptsize [10.79, 15.29]} & 12.29 {\scriptsize [10.07, 14.72]} \\
\quad 1b: other speakers only & 1,257 & 300 & \textbf{1.86} {\scriptsize [1.38, 2.44]} & 14.45 {\scriptsize [12.35, 16.77]} & 13.54 {\scriptsize [11.51, 15.74]} \\
2: known / new & 1,500 & 300 & \textbf{5.02} {\scriptsize [4.34, 5.78]} & 9.32 {\scriptsize [7.84, 11.01]} & 8.67 {\scriptsize [7.50, 9.95]} \\
3: new / known & 37,281 & 124 & \textbf{2.58} {\scriptsize [1.71, 3.60]} & 16.18 {\scriptsize [12.68, 20.03]} & 15.20 {\scriptsize [11.90, 18.86]} \\
4: new / new & 10,648 & 124 & \textbf{6.42} {\scriptsize [5.26, 7.96]} & 9.89 {\scriptsize [7.90, 12.52]} & 11.88 {\scriptsize [9.77, 14.49]} \\
\bottomrule
\end{tabular}}
\end{table}

Table~\ref{tab:cells} shows three observations.
\begin{itemize}\setlength\itemsep{0pt}
  \item \textbf{Familiar sentences were linked to a larger difference than familiar speakers.} For new speakers, CER
  is 2.58\% on known sentences against 6.42\% on new ones. A known speaker saying a new sentence has
  5.02\%. Known sentences are not easy in themselves: the original Parakeet model does \emph{worse} on them
  (16.18\%) than on new ones (9.89\%), so the low error comes from training on them. The groups also
  differ in speakers, prompt types and length, so this is an association, not a causal effect.
  \item \textbf{A person's own sentences had the lowest error rate.} CER was 0.75\% on held-back recordings of
  sentences the same speaker had said in training. For the 30 held-back speakers with the highest analysis-model
  CER, it was 5.84\% on such sentences against 29.03\% on new sentences; because these speakers were chosen
  by this model's own errors, the contrast is exaggerated.
  \item \textbf{Part of the gap is length.} Known prompts are often short commands. After matching recording
  lengths, known sentences still have about 1.7 times fewer errors than new ones (4.89\% vs
  8.32\%), down from about 2.5 times.
\end{itemize}
Evaluations that reuse training prompts may therefore show lower error rates than a system achieves on new text.

\subsection{Live transcription on CPUs}
\label{sec:stream}

Track~2 reports two latencies: the time until the first word appears and stays, and the time from the end of speech
to the final text. Our live entries run fine-tuned models of the same kind as our offline entries, on CPUs without a
GPU.

\begin{table}[t]
  \centering
  \caption{Our six scored Track~2 entries on Test1, as reported by the organisers: CER and the two median latencies in
  seconds. ``Sum'' adds the two medians.}
  \label{tab:track2}
  \small
  \begin{tabular}{lcccc}
\toprule
Track 2 entry & CER (\%) & First stable word (s) & Final text (s) & Sum (s) \\
\midrule
Entry 1 & 8.05 & 7.35 & 1.20 & 8.56 \\
Entry 2 & 7.98 & 0.82 & 1.51 & 2.34 \\
Entry 3 & 8.02 & 0.41 & 0.53 & 0.94 \\
Entry 4 & 7.96 & 0.41 & 0.52 & 0.93 \\
Entry 5 & 7.97 & 0.42 & 0.50 & 0.92 \\
Entry 6 & 8.07 & 0.42 & 0.44 & 0.85 \\
\bottomrule
\end{tabular}
\end{table}

In the challenge's CPU evaluation (Table~\ref{tab:track2}), the sum of the two median latencies fell from
8.6 seconds for our first entry to between 0.85 and 0.93 seconds for the later
ones. Live CER was 7.96--8.07\%, close to our offline entries (7.81--8.30\%). The fastest entry
was also the least accurate, so speed and accuracy still trade off slightly.

\subsection{Lessons for evaluating improvements}
\label{sec:negative}

We tested many other changes. Most did not show consistent improvements, and four general lessons stand out.
\begin{itemize}\setlength\itemsep{0pt}
  \item \textbf{Count speakers, not recordings.} On a 41-speaker panel, many measured gains were smaller than the
  speaker-level uncertainty, and those rarely held on other speakers or on Test1.
  \item \textbf{Check the hardest speakers separately.} Several changes that were neutral overall made results worse
  for slow or severely affected speakers.
  \item \textbf{Fix the rules before looking.} Setting pass/fail rules in advance, on panels separate from the
  selection panel, stopped us from adopting changes that only fit one panel.
  \item \textbf{Development panels are optimistic.} Test1 error was about 1.29 times selection-panel error,
  partly because a small panel has few of the hardest speakers.
\end{itemize}

\section{Implications}

\begin{itemize}\setlength\itemsep{0pt}
  \item \textbf{Fine-tuning substantially reduced character errors}, most of all for the hardest-to-understand
  speakers. Whether the remaining error rates are usable in practice is untested.
  \item \textbf{Personal recordings may help.} The low error rate on people's own repeated sentences suggests that
  recordings from the user could be valuable, in line with earlier work on personalised
  models~\citep{shor2019personalizing,green2021personalized,tobin2022personalized}. This is a hypothesis, not a result of
  this paper.
  \item \textbf{Report accuracy in parts.} Results by condition, by severity and by whether the sentence was known,
  with speaker-level uncertainty, are more informative than a single number.
\end{itemize}

\section{Limitations and ethics}

The selection panel has only 41 speakers (four with stroke), so per-condition intervals are wide. The severity and
familiarity analyses use a single fine-tuned model, not our best entry. Severity is measured by an off-the-shelf
recogniser's errors, not by a clinical rating. Known-speaker test recordings come from the same sessions as that
speaker's training recordings, which may overstate the speaker effect. The hidden test sets report only overall
scores, so we cannot break them down by condition.

All speech comes from the SAP corpus under its data use terms. We report only aggregate numbers: no audio, transcripts
or individual recordings appear here. Recognition is still much less accurate for some people, especially with severe
cerebral palsy, Down syndrome or stroke, so such systems should support, not replace, human communication partners,
and should be evaluated with the people who will use them.

\section{Conclusion}

Fine-tuning a public speech recogniser and averaging fine-tuned models roughly halved errors on our 41-speaker
selection panel and gave 7.81\% CER on Test1, against 15.26\% for the official baseline. In a
separate single-model analysis, the share of errors removed was largest for the hardest-to-understand speakers. Live
entries of the same kind ran on CPUs with about 0.9 seconds of combined median latency at similar accuracy. Having the
sentence in training was linked to a larger drop in errors than having the speaker in training, which matters for how
such systems are evaluated.

{\small
\bibliographystyle{plainnat}
\bibliography{refs}
}

\end{document}